\documentclass[a4paper,11pt]{article}

\usepackage{amssymb,amsmath,array}
\usepackage{hyperref}
\usepackage{bm}

\usepackage[dvips]{graphicx}
\usepackage{amssymb,amsmath,bm}
\usepackage{array,booktabs}
\newcolumntype{C}{>{\centering\arraybackslash}p{3cm}}

\usepackage{hyperref}

\usepackage[usenames,dvipsnames,svgnames]{xcolor}
\usepackage{tikz}
\usetikzlibrary{arrows}
\tikzstyle{edge}=[->, >=stealth', shorten <=2pt, shorten >=2pt, auto, line width=0.2mm]

\usepackage[a4paper, left=3cm, right=2.5cm, top=3cm, bottom=3cm]{geometry}

\usepackage{authblk}
\usepackage{natbib}
\usepackage{doi}

\usepackage[utf8]{inputenc}
\usepackage[T1]{fontenc}
\usepackage[english]{babel}
\usepackage{csquotes}

\renewcommand{\cite}{\citep}

\newcommand{\R}{\mathbb{R}}
\newcommand{\N}{\mathbb{N}}
\newcommand{\mat}[1]{\bm{#1}}
\newcommand{\transp}[1]{{#1}^T}

\newcommand{\graph}{\mathcal{G}}
\newcommand{\nodes}{V}
\newcommand{\edges}{E}
\newcommand{\lnode}{u}
\newcommand{\rnode}{v}

\newcommand{\lbl}{\xi}
\newcommand{\nodelim}{T}

\newcommand{\alphdims}{k}
\newcommand{\parents}{\mathcal{P}}
\newcommand{\children}{\mathcal{C}}
\newcommand{\neigh}{\mathcal{N}}

\newcommand{\tree}{\tilde x}
\newcommand{\seq}{\bar x}

\newcommand{\inspace}{\mathcal{X}}
\newcommand{\enc}{\phi}

\newcommand{\rec}{f}
\newcommand{\outfun}{g}
\newcommand{\dims}{n}
\newcommand{\point}{x}

\newcommand{\datalim}{m}

\newcommand{\kernel}{k}
\newcommand{\dist}{d}

\author{Benjamin Paaßen}
\affil{CITEC Center of Excellence, Bielefeld University\thanks{Funding by the CITEC center of excellence (EXC 277) is gratefully acknowledged.}}
\author{Claudio Gallicchio}
\author{Alessio Micheli}
\affil{Department of Computer Science, University of Pisa}
\author{Alessandro Sperduti}
\affil{Department of Mathematics, University of Padova}

\title{Embeddings and Representation Learning for Structured Data}

\date{Preprint of the ESANN 2019 paper \citet{Paassen2019ESANN2} as provided by the authors.
The original can be found at the \href{https://www.elen.ucl.ac.be/esann/proceedings/electronicproceedings.htm}{ESANN electronics proceedings page}.}

\begin{document}

\maketitle

\pagestyle{myheadings}
\markright{Preprint of \citet{Paassen2019ESANN2} provided by the authors.}

\nocite{Kriege2019,Navarin2019,Bakker2019,Mirus2019,Bacciu2019,Hautecoeur2019}

\begin{abstract}
Performing machine learning on structured data is complicated by the fact that such data
does not have vectorial form. Therefore, multiple approaches have emerged to construct vectorial
representations of structured data, from kernel and distance approaches to recurrent, recursive, and
convolutional neural networks. Recent years have seen heightened attention in this demanding field of
research and several new approaches have emerged, such as metric learning on structured data, graph
convolutional neural networks, and recurrent decoder networks for structured data. In this
contribution, we provide an high-level overview of the state-of-the-art in representation
learning and embeddings for structured data across a wide range of machine learning fields.
\end{abstract}

Traditional machine learning has mostly focused on the question of how to solve problems like
classification or regression for fixed, manually engineered data representations \cite{Bengio2013}.
By contrast, representation learning focuses on the challenge of obtaining a
vectorial representation in the first place, such that subsequent problems become easy to
solve \cite{Bengio2013}. Such an alternative view is particularly helpful for processing structured
data, i.e.\ sequences, trees, and graphs, where vectorial representations are not immediately available
\cite{Hamilton2017Review}.

A wide range of machine learning fields has attempted to construct such vectorial representations
for structured data.
In this contribution, we provide a high-level overview of these approaches, highlighting shared
foundations and properties. Thus, we hope to provide readers with a rich toolbox to handle structured
data and sufficient context knowledge to select the fitting method for any given situation.

We begin by introducing key concepts of structured data and vectorial representations before we
dive into the various approaches which have been proposed to achieve such representations.
This paper concludes with an overview of the contributions in this special session.

\section{Background}

In this section, we introduce terms that are shared among all methods for representation learning
and embeddings for structured data.
We begin by defining structured data itself. In particular, we define a graph as
a triple $\graph = (\nodes, \edges, \lbl)$, where $\nodes = \{\rnode_1, \ldots, \rnode_\nodelim\}$ is a finite set of nodes,
$\edges \subseteq \nodes \times \nodes$ is a finite set of edges, and $\lbl : \nodes \to \R^\alphdims$
is a mapping which assigns some vectorial label $\lbl(\rnode)$ to each node $\rnode \in \nodes$.

We call a node $\rnode$ a \emph{parent} of another node $\lnode$ if $(\rnode, \lnode) \in \edges$.
Conversely, we call $\lnode$ a \emph{child} of $\rnode$. We denote the set of all
parents of $\lnode$ as $\parents_\lnode$, the set of all children of $\rnode$ as $\children_\rnode$,
and we define the \emph{neighborhood} of $\lnode$ as $\neigh(\lnode) := \{\lnode\} \cup \parents_\lnode \cup \children_\lnode$.

We call a graph a \emph{tree} if exactly one node exists which has no parents (which we call the \emph{root})
and every other node has exactly one parent. We call all nodes without children \emph{leaves}.
As a notational shorthand, we also denote trees in a recursive fashion. In particular, if
$\lnode \in \nodes$ is the root of a tree and $\rnode_1, \ldots, \rnode_C$ are its children,
then the recursive tree notation is $\lnode(\tree_1, \ldots, \tree_C)$, where
$\tree_i$ is the recursive tree notation for the subtree rooted at $\rnode_i$.

We call a graph a \emph{sequence} if it is a tree with exactly one leaf or if it is the empty graph
$\epsilon = (\emptyset, \emptyset, \lbl)$. As a notational shorthand, we denote a sequence as
$\rnode_1, \ldots, \rnode_\nodelim$ where $(\rnode_t, \rnode_{t+1}) \in \edges$.

Now, we turn to embeddings and representations. Let $\inspace$ be some arbitrary set. Then, we call
a mapping $\enc : \inspace \to \R^\dims$ a $\dims$-dimensional \emph{embedding} of $\inspace$.
For any $\point \in \inspace$ we call $\enc(\point)$ the \emph{representation} of $\point$.

\section{Approaches for Embeddings of Structured Data}

\begin{table}
\setlength{\extrarowheight}{3pt}
\begin{center}
\begin{footnotesize}
\begin{tabular}{lCCCC}
& \textbf{sequences} & \textbf{trees} & \textbf{graphs} & \textbf{nodes} \\
\cmidrule(lr){1-1} \cmidrule(lr){2-4} \cmidrule(lr){5-5}
& \multicolumn{3}{c}{\textbf{kernels}} & \\
\cmidrule(lr){1-1} \cmidrule(lr){2-4} \cmidrule(lr){5-5}
fixed & n-grams, substrings~\cite{Lodhi2002} & subtrees~\cite{Aiolli2015}, reservoir activation~\cite{Bacciu2016}  & shortest paths \cite{Borgwardt2007}, Weisfeiler-Lehmann~\cite{Shervashidze2011}\textbf{\cite{Kriege2019}} & Laplacian~\cite{Kondor2016}, neighborhood hashing~\cite{Navarin2017} \\
learned & - & Markov models~\cite{Bacciu2018} & - & - \\
meta & \multicolumn{4}{>{\centering\arraybackslash}p{0.8\linewidth}}{multiple kernel learning~\cite{Goenen2011,Aiolli2015MKL}, \textbf{feature space composition \cite{Navarin2019}}} \\
\cmidrule(lr){1-1} \cmidrule(lr){2-4} \cmidrule(lr){5-5}
& \multicolumn{3}{c}{\textbf{distances}} & \\
\cmidrule(lr){1-1} \cmidrule(lr){2-4} \cmidrule(lr){5-5}
fixed & string edits~\cite{Levenshtein1965}, alignments~\cite{Vintsyuk1968,Gotoh1982} & tree edits~\cite{Zhang1989} & - & - \\
learned & string edits~\cite{Bellet2014}, \textbf{tf-idf \cite{Bakker2019}} & tree edits~\cite{Paassen2018ICML} & - & - \\
meta & \multicolumn{4}{>{\centering\arraybackslash}p{0.8\linewidth}}{dimensionality reduction~\cite{Sammon1969,Pekalska2005,VanDerMaaten2008}, multiple metric learning~\cite{Hosseini2015,Nebel2017}} \\
\cmidrule(lr){1-1} \cmidrule(lr){2-4} \cmidrule(lr){5-5}
& \multicolumn{3}{c}{\textbf{neural networks}} & \\
\cmidrule(lr){1-1} \cmidrule(lr){2-4} \cmidrule(lr){5-5}
encoding & echo state~\cite{Jaeger2004,Rodan2012,Gallicchio2016}, recurrent~\cite{Hochreiter1997,Cho2014,Greff2017}\textbf{\cite{Mirus2019}} & recursive~\cite{Sperduti1997,Hammer2002,Gallicchio2013} & recursive~\cite{Micheli2004,Hammer2005}, hierarchical convolutional~\cite{Ying2018} & recurrent \cite{Scarselli2009,Gallicchio2010,Li2016}, constructive \cite{Micheli2009}, convolutional~\cite{Kipf2017,Hamilton2017,Velickovic2018} \\
decoding & sequence to sequence~\cite{Sutskever2014,Xu2015} & (doubly) recurrent~\cite{Alvarez-Melis2017,Jin2018,Chen2018}, grammar-based~\cite{Kusner2017,Dai2018} & adjacency matrix~\cite{Simonovsky2018}, edge sequence~\cite{Liu2018,You2018} \textbf{\cite{Bacciu2019}} & - \\
\end{tabular}
\end{footnotesize}
\end{center}
\caption{An overview of the approaches surveyed in this paper. Methods are sorted into columns
according to the kind of structured data they process - either sequences, trees, graphs, or nodes within graphs.
Each block in the table marks a different class of method, either kernels, distances, or neural networks.
In kernels and distances, three rows distinguish between fixed representations, learned representations, and meta-representations
built on pre-existing representations. For neural networks, we distinguish
between networks that focus on encoding and networks which decode. Contributions of this special session
are highlighted via bold print.}
\label{tab:overview}
\end{table}

Approaches for embeddings of structured data can be distinguished along multiple axes.
For example, we can distinguish according to the kind of structured data that a method can process
- sequences, trees, or full graphs -, whether it generates an implicit or an explicit representation,
whether it generates fixed or learned representations, whether nodes or entire structures are embedded, and
whether decoding is possible or not.

We begin our list with \emph{kernels} and \emph{distances}, which compute pairwise measures
of proximity between structured data based on a pre-defined and fixed algorithm that implicitly
corresponds to a vectorial representation. By contrast, \emph{neural networks} learn explicit
vectorial representations which are learned from data. In particular, \emph{recurrent neural networks}
are designed to process sequential data, \emph{recursive neural networks} for tree-structured data,
and \emph{graph convolutional neural networks} for nodes of general graphs. However, there also
exist extensions to process nodes of graphs via recurrent neural networks or entire graphs via
recursive neural networks.

Note that all of these methods are initially limited to \emph{encoding} a structured datum into a vectorial
representation and can not \emph{decode} a vector back into structured data.
Our final section covers recent approaches to perform such decodings. A quick overview of  methods
surveyed in this paper is in Table~\ref{tab:overview}.

\paragraph{Kernels:}
We call a mapping $\kernel : \inspace \times \inspace \to \R$ over some set $\inspace$ a \emph{kernel}
iff an embedding $\enc : \inspace \to \R^\dims$ exists (for possibly infinite $\dims$), such that for all
$x, y \in \inspace$ it holds: $\kernel(x, y) = \transp{\enc(x)} \cdot \enc(y)$.
Therefore, every kernel on structured data relies - implicitly or explicitly - on an embedding
for structured data. In the past decade, a diverse range of structure kernels have emerged, but
the conceptual basis is typically the same. A structure kernel defines a class of $\datalim$ (possibly infinite)
characteristic substructures $\graph_1, \ldots, \graph_\datalim$
and defines the embedding $\enc$ as the sum of the embeddings for all these substructures.
More precisely, if $h : \inspace \to \N^\datalim$ is a mapping of structured data to histograms over
the selected substructures and $\rec : \{\graph_1, \ldots, \graph_\datalim\} \to \R^\dims$ is an embedding for the pre-defined
set of substructures, then the overall embedding $\enc$ is given as
$\enc(\graph) = \sum_{i=1}^\datalim h(\graph)_i \cdot \rec(\graph_i)$.
Examples of such substructures include string $n$-grams, substrings, random walks, shortest
paths, or subtrees \cite{Lodhi2002,Aiolli2015,Borgwardt2007,Shervashidze2011}.
Most recently, kernels have also been constructed based on reservoir activations for nodes \cite{Bacciu2016}  or  learned substructures, such as Markov Model hidden states for nodes
\cite{Bacciu2018}.
The embedding $\rec$ for the substructures can be as simple as mapping
the $i$th substructure to the $i$th unit vector, i.e.\ $\enc$ just counts the substructures.
Multiple works are specifically devoted to constructing node-specific kernels based on the graph
Laplacian or comparing node neighborhoods \cite{Kondor2016,Navarin2017}.

Note that, in most cases, it is infeasible to explicitly compute histograms over the substructures due to large or infinite $\datalim$.
Therefore, most kernels are computed directly as $\kernel(x, y) = \sum_{i=1}^\datalim \sum_{j=1}^\datalim h(x)_i \cdot h(y)_j \cdot \kernel'(\graph_i, \graph_j)$,
e.g.\ via some dynamic programming scheme \cite{Shervashidze2011}. As such, the embedding remains
implicit and can not be directly exploited for subsequent learning.

Instead of summing up embeddings of base kernels, one can also concatenate
such embeddings, which is the basis for \emph{multiple kernel learning} (MKL). 
Given a set of base embeddings for graphs $\rec_1, \ldots, \rec_\datalim$,
MKL learns factors $\alpha_1, \ldots, \alpha_\datalim \in \R^+$ for these embeddings
and defines the overall $\enc$ as $\enc(\graph)
= \big(\sqrt{\alpha_1} \cdot \rec_1(\graph), \ldots, \sqrt{\alpha_\datalim} \cdot \rec_\datalim(\graph)\big)$,
such that the resulting learned kernel is given as $\kernel(x, y) = \sum_{i=1}^m \alpha_i \cdot \transp{\rec_i(x)} \cdot \rec_i(y)$
\cite{Goenen2011,Aiolli2015MKL}.

\paragraph{Distances:}
Distance measures on structured data typically quantify distance in terms of effort that is needed
to transform one datum into another by means of discrete edit operations such as node
deletions, insertions, or replacements. This framework includes measures like the string edit distance, dynamic time warping,
alignment distances, or the tree edit distance \cite{Levenshtein1965,Vintsyuk1968,Gotoh1982,Zhang1989}.
These measures are all non-negative, self-equal, and symmetric, but do not necessarily conform to
the triangular inequality and are thus not necessarily proper metrics \cite{Nebel2017}.
As with kernels, distances are related to embeddings but are typically computed directly via
dynamic programming. In particular, it can be shown that for any self-equal and symmetric function
$\dist$ there exist two embeddings $\enc^+$ and $\enc^-$, such that for all $x, y \in \inspace$ it holds
$\dist(x, y)^2 = \lVert \enc^+(x) - \enc^+(y)\rVert^2 - \lVert \enc^-(x) - \enc^-(y)\rVert^2$ \cite{Pekalska2005}.
We can make this embedding explicit by dimensionality reduction methods such as
multi-dimensional scaling or t-SNE \cite{Sammon1969,VanDerMaaten2008}.

It is worth noting that edit distances can be learned in a supervised fashion by manipulating the
costs of single edits to facilitate classification \cite{Bellet2014,Paassen2018ICML}.

\tikzstyle{fun}=[circle,draw]
\tikzstyle{funin}=[edge]
\tikzstyle{funout}=[edge]

\begin{figure}
\begin{center}
\begin{tikzpicture}
\begin{scope}[shift={(-3.5,0)},scale=1.1]

\node[above] at (-0.5,0.3) {recurrent network};

\node (h0) at (0, 0) {$\enc(\epsilon)$};

\node[left] (x1) at (-1, -1) {$\lbl(a)$};
\node[fun] (f1) at (0, -1) {$\rec$};
\path[funin] (h0) edge (f1) (x1) edge (f1);

\node[left] (x2) at (-1, -2) {$\lbl(b)$};
\node[fun] (f2) at (0, -2) {$\rec$};
\path[funin] (f1) edge node[right] {$\enc(a)$} (f2) (x2) edge (f2);

\node[left] (x3) at (-1, -3) {$\lbl(c)$};
\node[fun] (f3) at (0, -3) {$\rec$};
\path[funin] (f2) edge node[right] {$\enc(a,b)$} (f3) (x3) edge (f3);
\node (h3) at (0, -4) {$\enc(a,b,c)$};
\path[funout] (f3) edge (h3);
\end{scope}

\begin{scope}[shift={(0,0)},scale=1.1]

\node[above] at (0,0.3) {recursive network};

\node (x1) at (-1, 0) {$\lbl(a)$};
\node[fun] (f1) at (-1,-1) {$\rec_a$};
\path[funin] (x1) edge (f1);

\node (x2) at (+1, 0) {$\lbl(b)$};
\node[fun] (f2) at (+1,-1) {$\rec_b$};
\path[funin] (x2) edge (f2);

\node[above] (x3) at (0, -1) {$\lbl(c)$};
\node[fun] (f3) at (0,-2) {$\rec_c$};
\node (h3) at (0,-3) {$\enc(c(a,b))$};
\path[funin] (x3) edge (f3)
(f1) edge node[below left] {$\enc(a)$} (f3)
(f2) edge node[below right] {$\enc(b)$} (f3);
\path[funout] (f3) edge (h3);
\end{scope}

\begin{scope}[shift={(4,0)},scale=1.1]

\node[above] at (0,0.25) {graph convolutional network};

\node (h0a) at (-1, 0) {$\lbl(a)$};
\node (h0b) at ( 0, 0) {$\lbl(b)$};
\node (h0c) at (+1, 0) {$\lbl(c)$};

\node[fun] (f1a) at (-1,-1.5) {$\rec^1$};
\path[funin] (h0a) edge (f1a);
\node[fun] (f1b) at ( 0,-1.5) {$\rec^1$};
\path[funin] (h0a) edge (f1b)
  (h0b) edge (f1b);
\node (p1c) at (+1,-0.6) {$+$};
\node[fun] (f1c) at (+1,-1.5) {$\rec^1$};
\path[funin] (h0a) edge[bend right] (p1c)
  (h0b) edge (p1c)
  (h0c) edge[bend left] (f1c)
  (p1c) edge[shorten <=0pt] (f1c);

\node[fun] (f2a) at (-1,-3) {$\rec^2$};
\path[funin] (f1a) edge node[left] {$\enc^1(a)$} (f2a);
\node[fun] (f2b) at ( 0,-3) {$\rec^2$};
\path[funin] (f1a) edge (f2b)
  (f1b) edge node[pos=0.8, right] {$\enc^1(b)$} (f2b);
\node (p2c) at (+1,-2.1) {$+$};
\node[fun] (f2c) at (+1,-3) {$\rec^2$};
\path[funin] (f1a) edge[bend right] (p2c)
  (f1b) edge (p2c)
  (f1c) edge[bend left] node[right] {$\enc^1(c)$} (f2c)
  (p2c) edge[shorten <=0pt] (f2c);

\node (h2a) at (-1, -4) {$\enc^2(a)$};
\node (h2b) at ( 0, -4) {$\enc^2(b)$};
\node (h2c) at (+1, -4) {$\enc^2(c)$};
\path[funout] (f2a) edge (h2a)
  (f2b) edge (h2b)
  (f2c) edge (h2c);

\end{scope}

\end{tikzpicture}
\end{center}
\caption{An illustration of the three encoder networks presented in the paper, namely a recurrent network which encodes
the sequence $a, b, c$ via Equation~\ref{eq:recurrent} (left), a recursive network that encodes the tree $c(a, b)$ via Equation~\ref{eq:recursive}
(center), and a two-layer graph convolutional neural network that encodes the nodes of the graph $\graph = (\{a, b, c\}, \{ (a, b), (a, c), (b, c) \})$
via Equation~\ref{eq:convolutional} (right).
Function applications/neurons are indicated by circles. The computational flow is indicated by arrows.}
\label{fig:encoders}
\end{figure}
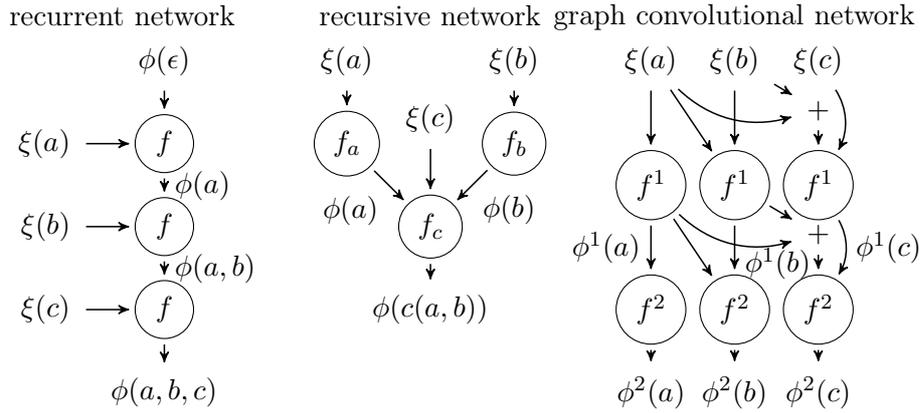

\paragraph{Recurrent Neural Networks:}
A recurrent neural network (RNN) maps sequential data $\seq = \vec \point_1, \ldots, \vec \point_\nodelim \in \R^\alphdims$
to a representation $\enc(\seq) \in \R^\dims$ by means of the recursive equation
\begin{equation}
\enc(\vec \point_1, \ldots, \vec \point_t) = \rec\big(\vec \point_t, \enc(\vec \point_1, \ldots, \vec \point_{t-1})\big), \label{eq:recurrent}
\end{equation}
where $\rec$ is some mapping $\rec : \R^\alphdims \times \R^\dims \to \R^\dims$ and where
$\rec(\epsilon)$ is typically defined as the zero vector (also refer to Figure~\ref{fig:encoders}, left).

In recurrent neural networks, the function $\rec$ is a neural network layer, e.g.\
a classic sigmoid layer of the form $\rec(\vec \point_t, \vec h_{t-1}) = \sigma\big( \mat U \cdot \vec \point_t + \mat W \cdot \vec h_{t-1})$,
where $\mat U \in \R^{\dims \times \alphdims}$ and $\mat W \in \R^{\dims \times \dims}$ are
weight matrices and $\sigma$ is a sigmoid function such as the tanh function.
A challenge in learning such networks are vanishing or exploding gradients over time,
which can be addressed, for instance, by the following strategies. First, one can decide to adapt neither
either $\mat U$ nor $\mat W$ but to initialize them in a randomized or deterministic fashion
\cite{Jaeger2004,Rodan2012}, also in a deep setting \cite{Gallicchio2016}. Second, one can replace a standard sigmoid layer
with a gated architecture that can ignore irrelevant parts of the sequence and thus maintain
memory over longer time without being unstable \cite{Hochreiter1997,Cho2014,Greff2017}.
For example, gated recurrent units \cite{Cho2014} define the recurrent function
$\rec(\vec \point_t, \vec h_{t-1}) = \vec z(\vec \point_t, \vec h_{t-1}) \odot \vec h_t
+ [\vec 1 - \vec z(\vec \point_t, \vec h_{t-1})] \odot \sigma\big( \mat U \cdot \vec \point_t + \mat W \cdot [\vec r(\vec \point_t, \vec h_{t-1}) \odot \vec h_{t-1}]\big)$,
where $\odot$ denotes the element-wise product and where $z(\vec \point_t, \vec h_{t-1}) \in [0, 1]^\dims$ as well as $\vec r(\vec \point_t, \vec h_{t-1}) \in [0, 1]^\dims$ are so-called gates,
computed via standard sigmoid layers as above.

The objective function of recurrent neural networks is typically to map the input sequence
$\seq$ to an output sequence $\bar y = \vec y_1, \ldots, \vec y_\nodelim \in \R^l$ by means of an output sigmoid layer
$\outfun : \R^\dims \to \R^l$ such that $\outfun(\vec h_t) \approx \vec y_t$ for all $t \in \{1, \ldots, \nodelim\}$.
However, recurrent neural networks can also be trained to auto-encode sequences or to decode
to other sequences \cite{Sutskever2014}.

Note that RNNs can also be applied to embed nodes in a graph by considering the embedding
$\enc(\rnode)$ of a node $\rnode \in \nodes$ as part of a state vector in a RNN \cite{Scarselli2009,Gallicchio2010,Li2016}. More precisely,
let $\enc^t(\rnode)$ denote the embedding of node $\rnode$ at time $t$. Then, we can
construct the node embedding at time $t+1$ via the equation $\enc^{t+1}(\rnode)
= \sum_{\lnode \in \neigh(\rnode)} \hat \rec\big(\lbl(\rnode), \enc^t(\rnode), \lbl(\lnode), \enc^t(\lnode)\big)$
or, similarly, via the equation $\enc^{t+1}(\rnode) = \hat \rec\Big(\lbl(\rnode), \enc^t(\rnode),
\sum_{\lnode \in \neigh(\rnode)} \lbl(\lnode), \sum_{\lnode \in \neigh(\rnode)} \enc^t(\lnode)\Big)$
for some mapping $\hat \rec : \R^\alphdims \times \R^\dims \times \R^\alphdims \times \R^\dims \to \R^\dims$.
In both cases, this collapses to a recurrent neural network according to Equation~\ref{eq:recurrent}
if we consider the concatenation of all node labels $\lbl(\rnode)$ as the input for each time step $t$
and the concatenation of all embeddings $\enc^t(\rnode)$ as state vector at time $t$.
The training of this network is typically guided by some supervised objective as in regular RNNs
\cite{Scarselli2009,Gallicchio2010,Li2016}.

As \cite{Scarselli2009} have shown, letting this network run is guaranteed to converge to a fix
point if $\hat \rec$ is a contractive map. In other words, one can let the network run for a
sufficiently large time and then use the resulting embedding at that time as an approximation of
the fix point and thus as an embedding of the nodes \cite{Li2016}.

\paragraph{Recursive Neural Networks:}
Recursive neural networks are an extension of recurrent neural networks for tree structured
data. Given a tree $\rnode(\tree_1, \ldots, \tree_C)$, a recursive neural network is defined by the
recursive equation
\begin{equation}
\enc\big(\rnode(\tree_1, \ldots, \tree_C)\big) = \rec_\rnode\big(\lbl(\rnode), \enc(\tree_1), \ldots, \enc(\tree_C)\big), \label{eq:recursive}
\end{equation}
where $\rec_\rnode$ is typically a sigmoid layer (also refer to Figure~\ref{fig:encoders}, center) \cite{Sperduti1997,Hammer2002,Gallicchio2013}.
In other words, the encoding starts with the leaves of the tree and then processes the tree bottom-up
until an embedding for the entire tree is obtained at the root.
Extensions of recursive neural networks to the treatment of  directed postional acyclic graphs
(DPAGs) have been introduced in \cite{Micheli2004,Hammer2005}.

Note that the construction of $\rec_\rnode$ by be challenging of the children have no clear
positional order or if the number of children is not consistent among nodes. Such problems can be
addressed by using order-invariant operators like sum or product to aggregate child embeddings,
to fill missing children with special tokens like zero vectors, to normalize the trees to binary
structure, or to learn specific functions for different kinds of nodes with different number
of children.

\paragraph{Graph Convolutional Neural Networks:}
Graph Convolutional Neural Networks (GCNs) generate embeddings of nodes in graphs similarly to
RNNs but via a layered feedforward architecture. In particular, let $\enc(\rnode)^l$ denote the
embedding of node $\rnode$ in layer $l$ of the network, where $\enc(\rnode)^0 = \lbl(\rnode)$.
Then, the embedding in layer $l+1$ is obtained via the equation
\begin{equation}
\enc(\rnode)^{l+1} = \rec^{l+1}\big(\enc(\rnode)^l, \sum_{\lnode \in \parents(\rnode)} \alpha(\lnode, \rnode) \cdot \enc(\lnode)^l\big), \label{eq:convolutional}
\end{equation}
where $\rec^l$ is a sigmoid layer and $\alpha(\lnode, \rnode)$ is some connectivity factor depending
on the graph structure (also refer to Figure~\ref{fig:encoders}, right) \cite{Kipf2017}.
Note that this equation differs from the RNN equation of \cite{Scarselli2009,Gallicchio2010,Li2016} in that we use different parameters
for each layer. Also note that the embedding in the $l$th layer integrates information from nodes up to distance $l$
in the graph. 

The idea to treat the mutual dependencies (graph cycles) through different neural network layers 
and to extend the nodes embedding by composition of the information of previous layers was originally
introduced (and formally proved) in the context of constructive approaches \cite{Micheli2009}.
Therein, the concept/terminology of visiting (the nodes) of the graphs corresponds to the
terminology of convolution over (the nodes) of the graphs used in GCN.
Indeed, the main differences between the model in \cite{Micheli2009} (NN4G) and GCN are related to
the use of an incremental construction of the deep NN for NN4G instead of the end-to-end training of
GCN (with advantage for NN4G in terms of divide et impera automatic design and layer by layer learning).
A recent model proposal using the costruction of NN4G in the context of generative models is in
\cite{Bacciu2018ICML}.

As with RNNs, GCNs are trained in a supervised fashion where the last layer of a GCN is considered
as the output of the network. Further, we can train GCNs semi-supervised by augmenting the
supervised loss with a term that forces neighboring nodes to have similar encodings \cite{Kipf2017}.
While vanilla GCNs are limited to graphs for which the structure is known a priori, multiple
authors have recently extended GCNs to unknown structure, either by normalizing the neighborhood
or by using attention mechanisms \cite{Hamilton2017,Velickovic2018}.

Importantly, the embeddings of GCNs can also be aggregated to achieve an embedding for the entire
graph by iteratively clustering nodes to coarser structures and aggregating the embeddings inside
structures by a pooling network \cite{Ying2018}.

\paragraph{Decoder Networks:}
Decoding vectorial representations back into structured da\-ta poses a significant challenge as
decoding trees or full graphs is provably harder compared to encoding them \cite{Hammer2002}.
Therefore, present decoding approaches focus on decoding sequential data via recurrent neural
networks.

Most prominently, sequence-to-sequence (seq2seq) learning first encodes a sequence as a vector
via a recurrent neural and then applies a second recurrent neural network which decodes the
sequence step by step until it returns a special end-of-sentence token \cite{Sutskever2014}.

Given the success of this scheme for hard machine learning tasks such as machine translation or
caption generation \cite{Sutskever2014,Xu2015}, researchers have also attempted to apply it to
trees or graphs by encoding these structures as sequences. In particular, we can re-write graphs
as a sequences of nodes and edges if we impose a order on the graph's nodes, e.g.\ via
breadth-first-search \cite{Liu2018,You2018}. After this re-representation, we can train a recurrent neural
network to generate one edge at a time and thus reconstruct the entire graph until an end-of-sentence
token is generated \cite{Liu2018,You2018}.

Alternatively, we can exploit grammatical knowledge about the domain. If our data can be described
by a context-free grammar (as in the case of chemical molecules or computer programs), generating
a structured datum reduces to a sequence of grammar rule applications. Therefore, we can train a
recurrent neural network which outputs the current sequence of grammar rules to decode a given
datum, which will then also be guaranteed to be syntactically correct
\cite{Alvarez-Melis2017,Jin2018,Kusner2017}. Indeed, grammatical structures can even
be used to impose semantic constraints, such as chemical bond properties \cite{Dai2018}.

\section{Special Session Contributions}

The contributions in this special session cover a wide range of approaches for representation
learning and embeddings for structured data, namely two kernel approaches \cite{Kriege2019,Navarin2019},
one distance approach \cite{Bakker2019}, one sequence encoding approach via recurrent neural networks \cite{Mirus2019},
one graph decoding approach via recurrent neural networks \cite{Bacciu2019}, and an extension
of non-negative matrix factorization to uncover structure in high-dimensional vectorial data \cite{Hautecoeur2019}
(also refer to Table~\ref{tab:overview}).

In more detail, \cite{Kriege2019} presents a variation of the Weisfeiler-Lehman graph kernel \cite{Shervashidze2011}
which combines the concept of optimal assignments with multiple kernel learning \cite{Aiolli2015MKL}.
In particular, they define the kernel between two graphs $\graph$ and $\graph'$ as
$\kernel(\graph, \graph') = \max_{M \subseteq \mathcal{B}(\graph, \graph')} \sum_{(\lnode, \rnode) \in M} \kernel'(\lnode, \rnode)$
where $\mathcal{B}(\graph, \graph')$ is the set of all possible bijections between the nodes of $\graph$ and $\graph'$
and where $\kernel'$ is a weighted Weisfeiler-Lehman base kernel over nodes. Recall that
the Weisfeiler-Lehman kernel counts subtree patterns in the neighborhood of a node. The
kernel variation employed by \cite{Kriege2019} applies weights to these subtree patterns and then
optimizes these weights via multiple kernel learning.

\cite{Navarin2019} propose a scheme which generates a more expressive
feature space from a base node kernel by means of a sum of outer products. In particular, the
encoding is defined as
$\enc(\graph) = \sum_{\rnode \in \nodes} \rec(\rnode) \cdot \sum_{i=1}^D \sum_{\lnode \in \neigh^i(\rnode)} \transp{\rec(\lnode)}$,
where $\rec$ is the embedding of the base node kernel and $\neigh^i(\rnode)$ is the
$i$-hop neighborhood of node $\rnode$. They also consider a version of this kernel
where features are selected based on their discriminative value in a linear classifier.

\cite{Bakker2019} perform metric learning to weigh text features obtained via tf-idf
and latent semantic analysis and obtain an explicit, low-dimensional embedding via
t-SNE \cite{VanDerMaaten2008}.

\cite{Mirus2019} encode a snapshot of a driving scene incorporating a variable number
of vehicles via the semantic pointer architecture \cite{Eliasmith2013} and encode
a sequence of such snapshots via long short-term memory networks \cite{Hochreiter1997,Greff2017}
in order to predict the future movement of a single vehicle.

With the the aim is to provide an adpative approach to graph generation from arbitrary distributions,
\cite{Bacciu2019} first represent graphs as sequences of edges, where the edges are
ordered according to their starting node, and auto-encode graphs as vectors by
means of a sequence-to-sequence \cite{Sutskever2014} network consisting of two gated
recurrent units \cite{Cho2014}, where the former encodes an edge sequence as a
vector and the second decodes the edge sequence from the code vector.

Finally, \cite{Hautecoeur2019} proposes a variation of hierarchical alternating
least squares (HALS) to infer non-negative polynomial signals which best explain
a given data set of sequences in the sense that the Euclidean distance between
the observed sequences and the sequences produced by a linear combination
of non-negative polynomials on the same interval is as small as possible.

Overall, the contributions in this special session push the boundaries of
embeddings for structured data forward across a wide range of approaches.
This reflects the more intense recent interest in such embeddings in the research
community overall and gives hope for further progress in the future.

\bibliography{literature} 
\bibliographystyle{plainnat}

\end{document}